\documentclass{article}

\usepackage[preprint]{neurips_2025}

\usepackage[utf8]{inputenc} 
\usepackage[T1]{fontenc}    
\usepackage{hyperref}       
\usepackage{url}            
\usepackage{graphicx}
\usepackage{booktabs}       
\usepackage{amsfonts}    
\usepackage{amsmath}
\usepackage{nicefrac}       
\usepackage{microtype}      
\usepackage{xcolor}         
\usepackage{booktabs}
\usepackage{makecell}

\title{One-Step Sampler for Boltzmann Distributions \\
via Drifting}

\author{%
Wenhan Cao\textsuperscript{1,*} \quad Keyu Yan\textsuperscript{1,*} \quad Lin Zhao\textsuperscript{1,\textdagger}
\\
\textsuperscript{1}Department of Electrical and Computer Engineering, National University of Singapore
\\
\texttt{\{wenhan,yky,zhaolin\}@nus.edu.sg}
\\
{\small \textsuperscript{*}Equal contribution \quad \textsuperscript{\textdagger}Corresponding author}
}

\begin{document}

\maketitle

\begin{abstract}
We present a drifting-based framework for amortized sampling of Boltzmann distributions defined by energy functions. The method trains a one-step neural generator by projecting samples along a Gaussian-smoothed score field from the current model distribution toward the target Boltzmann distribution. For targets specified only up to an unknown normalization constant, we derive a practical target-side drift from a smoothed energy and use two estimators: a local importance-sampling mean-shift estimator and a second-order curvature-corrected approximation. Combined with a mini-batch Gaussian mean-shift estimate of the sampler-side smoothed score, this yields a simple stop-gradient objective for stable one-step training. On a four-mode Gaussian-mixture Boltzmann target, our sampler achieves mean error $0.0754$, covariance error $0.0425$, and RBF MMD $0.0020$. Additional double-well and banana targets show that the same formulation also handles nonconvex and curved low-energy geometries. Overall, the results support drifting as an effective way to amortize iterative sampling from Boltzmann distributions into a single forward pass at test time.
\end{abstract}

\section{Introduction}

Sampling from complex, high-dimensional distributions underlies many important problems in computational science, with applications spanning molecular modeling, Bayesian inference, and generative modeling. In this paper, we are interested in sampling from target distributions specified through an energy function $E(x)$, namely Boltzmann distributions of the form
\begin{equation}
p(x) = \frac{1}{Z}\exp(-E(x)),
\nonumber
\end{equation}
where
\begin{equation}
Z = \int \exp(-E(x))\,dx < \infty
\nonumber
\end{equation}
is an unknown normalization constant. More generally, one may write
\begin{equation}
p(x)=\frac{1}{Z}\exp\!\left(-\frac{1}{\tau}E(x)\right),
\nonumber
\end{equation}
with temperature $\tau>0$; throughout this paper, we absorb $\tau$ into $E$ and set $\tau=1$ for simplicity.

Boltzmann distributions provide a flexible way to represent structured target densities through their energy landscapes. Their main difficulty, however, lies in sampling. Since the normalization constant is typically intractable, exact sampling is unavailable, and one usually resorts to iterative procedures such as Langevin dynamics or Hamiltonian Monte Carlo~\citep{roberts1996exponential,roberts1998optimal,duane1987hybrid,neal2011mcmc}. These samplers often require many target evaluations and can mix poorly in complex or multi-modal landscapes, making them expensive to use at test time.

A natural alternative is to learn an amortized sampler~\citep{gershman2014amortized,kingma2014auto,rezende2015variational,dinh2017density}: a neural generator
\begin{equation}
x = f_\theta(z), \qquad z \sim p_0,
\nonumber
\end{equation}
whose pushforward distribution $q_\theta$ approximates the target $p$. Once trained, such a sampler produces approximate draws from the Boltzmann distribution using a single forward pass. The challenge is how to train $f_\theta$ when the target is specified only up to its energy function and an unknown normalization constant.

In this paper, we propose to use \emph{drifting model} \citep{deng2026generative} as a general framework for training neural samplers for Boltzmann distributions, closely related to particle transport ideas such as Stein variational gradient descent~\citep{liu2016stein}. Drifting casts sampler learning as a projected transport problem: given the current sampler distribution $q_\theta$, one defines a transport field that moves samples from $q_\theta$ toward the target $p$, and then regresses the generator toward the transported samples. This perspective is especially appealing here because the target-side transport can be expressed directly in terms of the energy function, without requiring the normalization constant.

Our starting point is to instantiate drifting with the Gaussian-smoothed score operator~\citep{hyvarinen2005estimation,hyvarinen2007extensions,vincent2011connection,song2021score}. Under this choice, the drift field is given by the difference between the smoothed target score and the smoothed sampler score. For a Boltzmann target, we show that the smoothed score admits a convenient representation in terms of a smoothed energy, and can be interpreted either as a local posterior mean-shift or as a locally averaged energy gradient. This leads to a practical training objective for neural samplers of Boltzmann distributions.

We further develop two approximations for the target-side drift. The first is a local importance-sampling estimator based on Gaussian perturbations around the current sample. The second is a second-order approximation that yields a curvature-corrected energy gradient, in spirit related to modern score-based and diffusion modeling developments~\citep{sohl2015deep,ho2020denoising,song2019sliced}. Combined with a simple mini-batch estimator of the sampler-side smoothed score, these approximations result in an efficient algorithm for amortized sampling from Boltzmann targets.
 
\paragraph{Contributions.}
Our contributions are threefold. First, we formulate sampling from Boltzmann distributions as a drifting problem and derive a training objective based on smoothed-score transport. Second, we characterize the target-side drift induced by an energy function and show how it can be approximated efficiently in practice. Third, we obtain a simple stop-gradient algorithm that trains a neural sampler to approximate Boltzmann sampling dynamics in an amortized manner.

\section{Preliminaries}

\subsection{Drifting}

Let $p$ denote the target distribution and let $q_\theta$ be the distribution induced by a generator
\begin{equation}
x = f_\theta(z), \qquad z \sim p_0.
\nonumber
\end{equation}
Drifting introduces a vector field $V_{p,q_\theta}(x)$ that specifies how a sample $x \sim q_\theta$ should be moved so as to better match the target distribution. The corresponding training objective is
\begin{equation}
\mathcal{L}_{\mathrm{Drift}}(\theta)
=
\mathbb{E}_{z \sim p_0}
\left\|
f_\theta(z)
-
\mathrm{sg}\left(
f_\theta(z)+V_{p,q_\theta}(f_\theta(z))
\right)
\right\|_2^2,
\nonumber
\end{equation}
where $\mathrm{sg}(\cdot)$ denotes the stop-gradient operator. At the objective level, this is equivalent to
\begin{equation}
\mathcal{L}_{\mathrm{Drift}}(\theta)
=
\mathbb{E}_{x \sim q_\theta}
\|V_{p,q_\theta}(x)\|_2^2.
\nonumber
\end{equation}

The vector field should vanish when the current sampler already matches the target. A sufficient condition is anti-symmetry,
\begin{equation}
V_{p,q}(x) = -V_{q,p}(x),
\nonumber
\end{equation}
which implies $V_{p,p}(x)=0$. A natural construction is
\begin{equation}
V_{p,q}(x)
=
\eta \bigl(T[p](x)-T[q](x)\bigr),
\nonumber
\end{equation}
where $T$ is an operator on distributions and $\eta>0$ is a step size.

\subsection{Smoothed-score drifting}

In this work, we choose $T$ to be the Gaussian-smoothed score operator,
\begin{equation}
T_\sigma[p](x)
=
\nabla_x \log (p * \phi_\sigma)(x),
\nonumber
\end{equation}
where $\phi_\sigma$ is the Gaussian kernel with bandwidth $\sigma$. The resulting drift is
\begin{equation}
V_{p,q}^{(\sigma)}(x)
=
\eta
\Bigl(
\nabla_x \log (p * \phi_\sigma)(x)
-
\nabla_x \log (q * \phi_\sigma)(x)
\Bigr).
\nonumber
\end{equation}
The drifting objective then becomes
\begin{equation}
\mathcal{L}_{\mathrm{Drift}}(\theta)
=
\eta^2
\mathbb{E}_{x \sim q_\theta}
\left\|
\nabla_x \log (p * \phi_\sigma)(x)
-
\nabla_x \log (q_\theta * \phi_\sigma)(x)
\right\|_2^2.
\nonumber
\end{equation}
This can be viewed as a reverse Fisher-type discrepancy between the smoothed target and sampler scores.

\subsection{Problem setup: amortized sampling for Boltzmann distributions}

We focus on the case where the target is a Boltzmann distribution. Classical training strategies for such energy-defined models include contrastive divergence and noise-contrastive estimation~\citep{hinton2002training,gutmann2010noise}, and modern variants include cooperative and short-run training schemes~\citep{xie2016cooperative,han2017alternating,nijkamp2019anatomy}. Here, \emph{amortized sampling} means learning a parametric sampler once during training so that test-time sampling is obtained by a single forward pass, thereby amortizing the cost of iterative per-sample MCMC updates.
\begin{equation}
p(x)=\frac{1}{Z}\exp(-E(x)),
\nonumber
\end{equation}
with energy function $E(x)$ and unknown partition function $Z$. Our goal is to train a generator $f_\theta$ such that $q_\theta \approx p$, thereby replacing iterative test-time MCMC with a single forward pass through the generator.

The central technical problem is therefore to compute or approximate the target-side smoothed score
\begin{equation}
\nabla_x \log (p * \phi_\sigma)(x)
\nonumber
\end{equation}
directly from the energy function.

\section{Method}

\subsection{Target-side drift for Boltzmann distributions}

For a Boltzmann target
\begin{equation}
p(x)=\frac{1}{Z}\exp(-E(x)),
\nonumber
\end{equation}
define the Gaussian-smoothed density
\begin{equation}
p_\sigma(x)
:=
(p * \phi_\sigma)(x)
=
\frac{1}{Z}
\int
\exp(-E(u)) \phi_\sigma(x-u)\,du.
\nonumber
\end{equation}
For a Gaussian kernel,
\begin{equation}
\phi_\sigma(x-u)
=
(2\pi\sigma^2)^{-d/2}
\exp\left(
-\frac{\|u-x\|^2}{2\sigma^2}
\right),
\nonumber
\end{equation}
we can rewrite $\log p_\sigma(x)$, up to an additive constant independent of $x$, as
\begin{equation}
\log p_\sigma(x)
=
-\bar E_\sigma(x)+C,
\nonumber
\end{equation}
where
\begin{equation}
\bar E_\sigma(x)
:=
-\log
\int
\exp\left(
-E(u)-\frac{\|u-x\|^2}{2\sigma^2}
\right)\,du.
\nonumber
\end{equation}
It follows that
\begin{equation}
\nabla_x \log (p * \phi_\sigma)(x)
=
-\nabla_x \bar E_\sigma(x).
\nonumber
\end{equation}
Hence the Boltzmann drift takes the form
\begin{equation}
V_{E,q}^{(\sigma)}(x)
=
\eta
\Bigl(
-\nabla_x \bar E_\sigma(x)
-
\nabla_x \log (q * \phi_\sigma)(x)
\Bigr),
\nonumber
\end{equation}
and the corresponding objective is
\begin{equation}
\mathcal{L}_{E\text{-}\mathrm{Drift}}(\theta)
=
\eta^2
\mathbb{E}_{x \sim q_\theta}
\left\|
-\nabla_x \bar E_\sigma(x)
-
\nabla_x \log (q_\theta * \phi_\sigma)(x)
\right\|_2^2.
\nonumber
\end{equation}

This shows that sampling from Boltzmann distributions can be cast as matching the sampler-side smoothed score to an energy-induced target-side drift.

\subsection{Interpreting the smoothed energy gradient}

To obtain a more explicit form, define
\begin{equation}
Z_\sigma(x)
:=
\int
\exp\left(
-E(u)-\frac{\|u-x\|^2}{2\sigma^2}
\right)\,du,
\nonumber
\end{equation}
so that $\bar E_\sigma(x)=-\log Z_\sigma(x)$. Differentiating yields
\begin{equation}
-\nabla_x \bar E_\sigma(x)
=
\frac{1}{\sigma^2}
\left(
\mathbb{E}_{u\sim\pi_\sigma(\cdot|x)}[u]-x
\right),
\nonumber
\end{equation}
where
\begin{equation}
\pi_\sigma(u|x)
:=
\frac{
\exp\left(
-E(u)-\frac{\|u-x\|^2}{2\sigma^2}
\right)
}{
Z_\sigma(x)
}
\nonumber
\end{equation}
is a local posterior centered at $x$.

Equivalently, by integration by parts,
\begin{equation}
-\nabla_x \bar E_\sigma(x)
=
-\mathbb{E}_{u\sim\pi_\sigma(\cdot|x)}
\bigl[\nabla_u E(u)\bigr].
\nonumber
\end{equation}
Therefore, the target-side drift can be interpreted either as a local mean-shift under the energy model or as a locally averaged energy gradient. This characterization is useful because it turns the intractable global sampling problem into a local estimation problem around each current sample.

\subsection{Approximating the target-side drift}

\paragraph{Monte Carlo approximation.}
A simple estimator can be obtained from local Gaussian perturbations. For a given sample $x$, draw
\begin{equation}
u_\ell = x + \sigma \varepsilon_\ell,
\qquad
\varepsilon_\ell \sim \mathcal{N}(0,I),
\qquad \ell=1,\dots,L,
\nonumber
\end{equation}
and define
\begin{equation}
w_\ell = \exp(-E(u_\ell)),
\qquad
\bar w_\ell = \frac{w_\ell}{\sum_{m=1}^L w_m}.
\nonumber
\end{equation}
Then
\begin{equation}
-\nabla_x \bar E_\sigma(x)
\approx
\frac{1}{\sigma^2}
\left(
\sum_{\ell=1}^L \bar w_\ell u_\ell - x
\right).
\nonumber
\end{equation}
This estimator performs a local importance-weighted mean-shift toward low-energy regions.

\paragraph{Second-order approximation.}
When second-order information is available, we can derive a local closed-form approximation. Let
\begin{equation}
g(x)=\nabla E(x), \qquad H(x)=\nabla^2 E(x).
\nonumber
\end{equation}
Using the quadratic expansion
\begin{equation}
E(u)
\approx
E(x)
+
g(x)^\top (u-x)
+
\frac{1}{2}(u-x)^\top H(x)(u-x),
\nonumber
\end{equation}
the local posterior becomes approximately Gaussian, yielding
\begin{equation}
\mathbb{E}[u-x \mid x]
\approx
-\sigma^2 \bigl(I+\sigma^2 H(x)\bigr)^{-1} g(x).
\nonumber
\end{equation}
Therefore,
\begin{equation}
-\nabla_x \bar E_\sigma(x)
\approx
-\bigl(I+\sigma^2 H(x)\bigr)^{-1}\nabla E(x).
\nonumber
\end{equation}
This approximation reveals that the smoothed target score is a curvature-corrected energy gradient.

\subsection{Estimating the sampler-side smoothed score}

The second term in the drift,
\begin{equation}
\nabla_x \log(q_\theta * \phi_\sigma)(x),
\nonumber
\end{equation}
depends on the current sampler distribution. We estimate it directly from a mini-batch using Gaussian mean-shift. Given generated samples $\{x_j\}_{j=1}^N$, define
\begin{equation}
K_\sigma(x_j,x_i)
=
\exp\left(
-\frac{\|x_j-x_i\|^2}{2\sigma^2}
\right).
\nonumber
\end{equation}
The sampler-side smoothed score at $x_i$ is estimated by
\begin{equation}
\hat s_{q,\sigma}(x_i)
=
\frac{1}{\sigma^2}
\left(
\frac{\sum_{j=1}^N K_\sigma(x_j,x_i)x_j}
{\sum_{j=1}^N K_\sigma(x_j,x_i)}
-
x_i
\right).
\nonumber
\end{equation}

\subsection{Training objective}

In practice, we optimize the stop-gradient form of drifting rather than directly differentiating the squared field norm. For a latent sample $z\sim p_0$, let
\begin{equation}
x=f_\theta(z).
\nonumber
\end{equation}
We first estimate the drift
\begin{equation}
\hat V(x)
=
\eta \bigl( \hat g_{E,\sigma}(x)-\hat s_{q,\sigma}(x) \bigr),
\nonumber
\end{equation}
where $\hat g_{E,\sigma}(x)$ is computed using either the Monte Carlo or second-order approximation. We then construct the frozen target
\begin{equation}
\tilde x = \mathrm{sg}\bigl(x+\hat V(x)\bigr),
\nonumber
\end{equation}
and minimize
\begin{equation}
\mathcal{L}_{E\text{-}\mathrm{Drift}}^{\mathrm{sg}}(\theta)
=
\mathbb{E}_{z \sim p_0}
\|f_\theta(z)-\tilde x\|_2^2.
\nonumber
\end{equation}

This objective has a simple interpretation: samples produced by the current generator are first transported toward the Boltzmann target under the estimated drift, and the generator is then updated to match the transported samples. In this way, the generator amortizes the local transport dynamics induced by the energy model.

\subsection{Mini-batch implementation}

Given a mini-batch
\begin{equation}
z_i \sim p_0, \qquad x_i=f_\theta(z_i), \qquad i=1,\dots,N,
\nonumber
\end{equation}
our algorithm proceeds as follows:
\begin{enumerate}
    \item Estimate the sampler-side smoothed score $\hat s_{q,\sigma}(x_i)$ using Gaussian mean-shift over the current batch.
    \item Estimate the target-side drift $\hat g_{E,\sigma}(x_i)$ using either local importance sampling or the second-order approximation.
    \item Form the drift
    \begin{equation}
    \hat V_i
    =
    \eta \bigl(\hat g_{E,\sigma}(x_i)-\hat s_{q,\sigma}(x_i)\bigr).
    \nonumber
    \end{equation}
    \item Construct frozen targets
    \begin{equation}
    \tilde x_i = \mathrm{sg}(x_i+\hat V_i).
    \nonumber
    \end{equation}
    \item Update $\theta$ by minimizing
    \begin{equation}
    \hat{\mathcal{L}}_{E\text{-}\mathrm{Drift}}^{\mathrm{sg}}(\theta)
    =
    \frac{1}{N}\sum_{i=1}^N \|x_i-\tilde x_i\|_2^2.
    \nonumber
    \end{equation}
\end{enumerate}

\section{Experiments}

\subsection{Experimental setup}

We evaluate Gaussian kernel drifting on the two-dimensional Gaussian-mixture Boltzmann distribution used throughout the paper. The target distribution has four symmetric modes, so the experiment tests whether a one-step amortized sampler can recover both the local shape of each mode and the global mode balance. In this draft, we focus on understanding the behavior of our sampler itself rather than comparing against external baselines.

For Gaussian kernel drifting, we use a residual MLP generator with latent dimension $32$, hidden width $256$, and three hidden layers. The model is trained for $10{,}000$ drifting updates with batch size $1024$ and learning rate $10^{-3}$. The Gaussian kernel bandwidth and drift step are both set to $\sigma=\eta=0.22$, and the target-side drift is estimated by the Monte Carlo local-mean-shift estimator with $256$ local perturbations per sample.

All reported numbers use $5000$ generated samples and $5000$ reference samples from the target, with random seed $42$. We report the $\ell_2$ error of the sample mean, the Frobenius norm of the covariance error, RBF MMD, and the mean energy of generated samples. Lower is better for the first three metrics, while the mean energy should be interpreted relative to the reference energy.

\subsection{Quantitative summary}

Table~\ref{tab:gaussian-kernel-summary} summarizes the final sampling quality of the trained one-step sampler. The learned generator attains mean error $0.0754$, covariance error $0.0425$, and RBF MMD $0.0020$, indicating that the generated distribution closely matches the target in both first- and second-order structure. The generated mean energy $1.0045$ is also close to the reference value $1.0263$, which shows that the transported samples lie in the correct low-energy region of the Boltzmann target.

\begin{table}[t]
\centering
\small
\begin{tabular}{lccccc}
\toprule
Method & Mean $\ell_2 \downarrow$ & Cov. Fro. $\downarrow$ & RBF MMD $\downarrow$ & Generated energy & Reference energy \\
\midrule
Gaussian kernel drifting & 0.0754 & 0.0425 & 0.0020 & 1.0045 & 1.0263 \\
\bottomrule
\end{tabular}
\caption{Quantitative summary for Gaussian kernel drifting on the Gaussian-mixture Boltzmann target. Lower is better for mean error, covariance error, and MMD.}
\label{tab:gaussian-kernel-summary}
\end{table}

Taken together, these numbers suggest that the drifting objective is sufficient to train a single neural map whose pushforward already matches the target distribution well, without any iterative sampling procedure at test time.

\subsection{Qualitative analysis}

\begin{figure}[t]
\centering
\includegraphics[width=\linewidth]{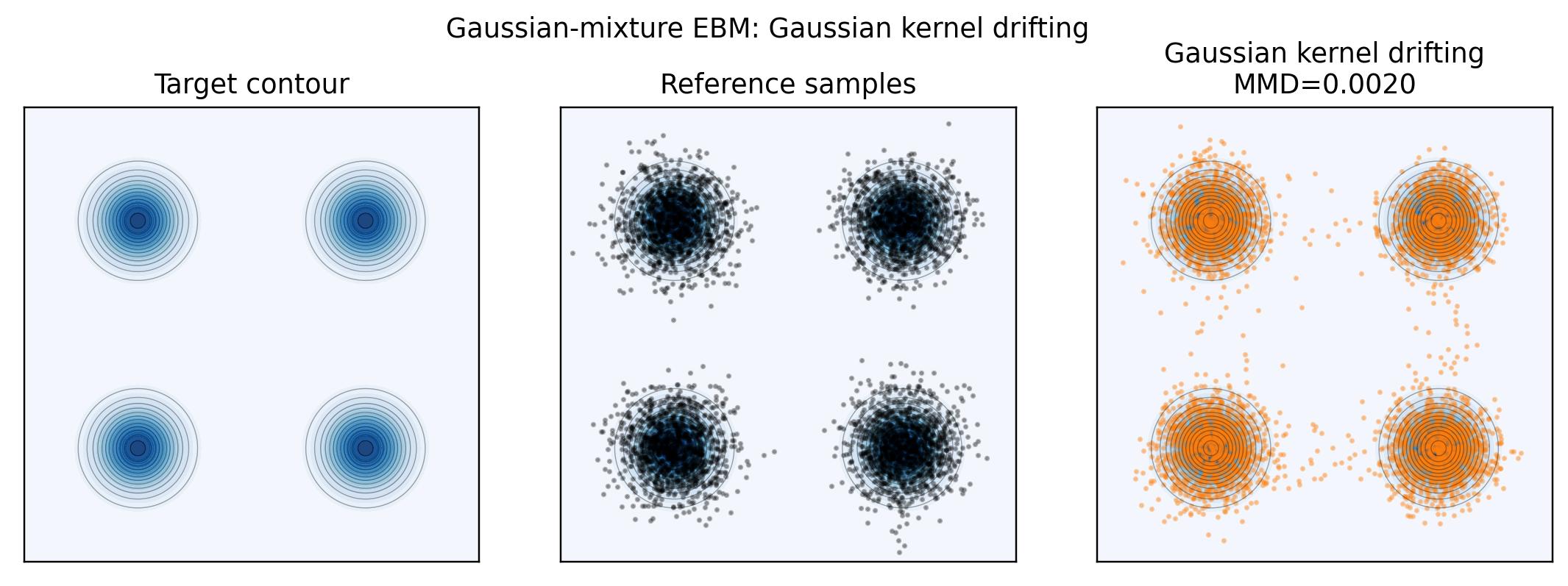}
\caption{Qualitative behavior of Gaussian kernel drifting on the Gaussian-mixture Boltzmann target. The learned one-step sampler captures all four modes at the correct locations and with approximately the correct spread.}
\label{fig:gaussian-kernel-qualitative}
\end{figure}

Figure~\ref{fig:gaussian-kernel-qualitative} visualizes the learned sampler together with the target contour and reference samples. The generated samples clearly occupy all four modes and align well with the target geometry. The mass allocation across the four quadrants is $(1141, 1244, 1332, 1283)$ out of $5000$ samples, which is close to the ideal balanced count of $1250$ per mode for this symmetric target.

The figure also reveals the main residual error mode of the sampler: a small amount of probability mass appears between neighboring modes, especially along the lower half of the square. This is consistent with the fact that the generator is trained as a smooth one-step transport map. Even so, the learned samples remain concentrated near the correct basins, and the qualitative picture matches the low MMD and covariance errors reported in Table~\ref{tab:gaussian-kernel-summary}.

\begin{figure}[t]
\centering
\includegraphics[width=0.9\linewidth]{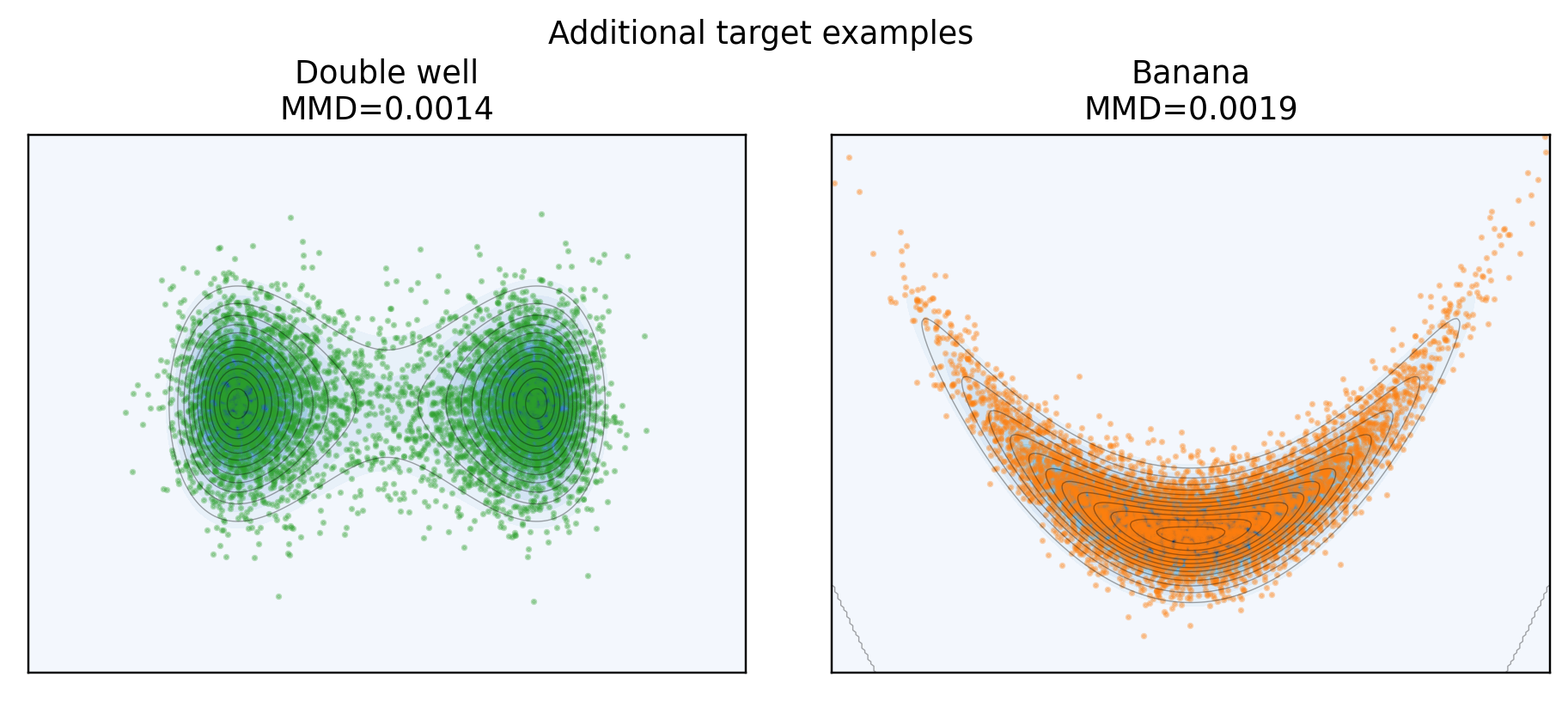}
\caption{Additional target examples for Gaussian kernel drifting. On the left, the learned sampler captures the two-well structure of the double-well energy. On the right, it follows the curved low-energy manifold of the banana-shaped target.}
\label{fig:additional-targets}
\end{figure}

To illustrate that the same procedure extends beyond the Gaussian-mixture example, Figure~\ref{fig:additional-targets} shows results on the double-well and banana targets with the same Gaussian kernel drifting framework. In the double-well case, the sampler recovers the two low-energy basins and the connecting bridge region, achieving mean error $0.0308$, covariance error $0.0326$, and MMD $0.0014$. In the banana case, the generated samples follow the nonlinear curved geometry of the target, with mean error $0.0226$ and MMD $0.0019$. These additional examples suggest that the drifting objective is not restricted to mixtures of nearly Gaussian modes, but also adapts to targets with nonconvex and highly curved support.

\section{Conclusion}

We proposed a drifting-based approach to amortized sampling for Boltzmann distributions, where a one-step generator is trained to match transported samples under a Gaussian-smoothed score field. The framework gives a concrete target-side drift for unnormalized energies and supports both Monte Carlo local mean-shift and second-order approximations, together with a mini-batch estimator for the sampler-side score. This combination leads to a practical and lightweight training algorithm.

Empirically, the method learns high-quality one-step samplers on multimodal toy Boltzmann distributions, with strong quantitative agreement on moment errors and MMD and with good qualitative mode coverage. Results on Gaussian-mixture, double-well, and banana targets suggest that the approach is robust beyond near-Gaussian settings and can follow curved low-energy manifolds.

Future work will test the method on higher-dimensional image Boltzmann distributions, compare directly to strong iterative samplers under matched compute budgets, and improve stability via adaptive bandwidths and lower-variance target-side estimators.
\bibliographystyle{plainnat}
\bibliography{references}

@article{deng2026generative,
  title={Generative Modeling via Drifting},
  author={Deng, Mingyang and Li, He and Li, Tianhong and Du, Yilun and He, Kaiming},
  journal={arXiv preprint arXiv:2602.04770},
  year={2026}
}

@article{hinton2002training,
  title={Training products of experts by minimizing contrastive divergence},
  author={Hinton, Geoffrey E},
  journal={Neural computation},
  volume={14},
  number={8},
  pages={1771--1800},
  year={2002},
  publisher={MIT Press}
}

@inproceedings{gutmann2010noise,
  title={Noise-contrastive estimation: A new estimation principle for unnormalized statistical models},
  author={Gutmann, Michael and Hyv{\"a}rinen, Aapo},
  booktitle={Proceedings of the thirteenth international conference on artificial intelligence and statistics},
  pages={297--304},
  year={2010},
  organization={JMLR Workshop and Conference Proceedings}
}

@article{hyvarinen2005estimation,
  title={Estimation of non-normalized statistical models by score matching.},
  author={Hyv{\"a}rinen, Aapo and Dayan, Peter},
  journal={Journal of Machine Learning Research},
  volume={6},
  number={4},
  year={2005}
}

@article{hyvarinen2007extensions,
  title={Some extensions of score matching},
  author={Hyv{\"a}rinen, Aapo},
  journal={Computational Statistics \& Data Analysis},
  volume={51},
  number={5},
  pages={2499--2512},
  year={2007}
}

@article{vincent2011connection,
  title={A connection between score matching and denoising autoencoders},
  author={Vincent, Pascal},
  journal={Neural Computation},
  volume={23},
  number={7},
  pages={1661--1674},
  year={2011}
}

@inproceedings{song2019sliced,
  title={Sliced score matching: A scalable approach to density and score estimation},
  author={Song, Yang and Garg, Sahaj and Shi, Jiaxin and Ermon, Stefano},
  booktitle={Proceedings of The 35th Uncertainty in Artificial Intelligence Conference},
  series={Proceedings of Machine Learning Research},
  volume={115},
  pages={574--584},
  year={2020}
}

@inproceedings{song2021score,
  title={Score-based generative modeling through stochastic differential equations},
  author={Song, Yang and Sohl-Dickstein, Jascha and Kingma, Diederik P. and Kumar, Abhishek and Ermon, Stefano and Poole, Ben},
  booktitle={International Conference on Learning Representations},
  year={2021}
}

@inproceedings{sohl2015deep,
  title={Deep unsupervised learning using nonequilibrium thermodynamics},
  author={Sohl-Dickstein, Jascha and Weiss, Eric and Maheswaranathan, Niru and Ganguli, Surya},
  booktitle={International Conference on Machine Learning},
  pages={2256--2265},
  year={2015}
}

@inproceedings{ho2020denoising,
  title={Denoising diffusion probabilistic models},
  author={Ho, Jonathan and Jain, Ajay and Abbeel, Pieter},
  booktitle={Advances in Neural Information Processing Systems},
  year={2020}
}

@article{duane1987hybrid,
  title={Hybrid Monte Carlo},
  author={Duane, Simon and Kennedy, Anthony D. and Pendleton, Brian J. and Roweth, Duncan},
  journal={Physics Letters B},
  volume={195},
  number={2},
  pages={216--222},
  year={1987}
}

@article{neal2011mcmc,
  title={MCMC using Hamiltonian dynamics},
  author={Neal, Radford M},
  journal={Handbook of markov chain monte carlo},
  pages={47--95},
  year={2011},
  publisher={Chapman and Hall/CRC}
}

@article{roberts1996exponential,
  title={Exponential convergence of Langevin distributions and their discrete approximations},
  author={Roberts, Gareth O. and Tweedie, Richard L.},
  journal={Bernoulli},
  volume={2},
  number={4},
  pages={341--363},
  year={1996}
}

@article{roberts1998optimal,
  title={Optimal scaling of discrete approximations to Langevin diffusions},
  author={Roberts, Gareth O. and Rosenthal, Jeffrey S.},
  journal={Journal of the Royal Statistical Society: Series B},
  volume={60},
  number={1},
  pages={255--268},
  year={1998}
}

@inproceedings{liu2016stein,
  title={Stein variational gradient descent: A general purpose Bayesian inference algorithm},
  author={Liu, Qiang and Wang, Dilin},
  booktitle={Advances in Neural Information Processing Systems},
  year={2016}
}

@inproceedings{gershman2014amortized,
  title={Amortized inference in probabilistic reasoning},
  author={Gershman, Samuel J. and Goodman, Noah D.},
  booktitle={Proceedings of the Annual Meeting of the Cognitive Science Society},
  volume={36},
  number={36},
  pages={517--522},
  year={2014}
}

@inproceedings{kingma2014auto,
  title={Auto-encoding variational Bayes},
  author={Kingma, Diederik P. and Welling, Max},
  booktitle={International Conference on Learning Representations},
  year={2014}
}

@inproceedings{rezende2015variational,
  title={Variational inference with normalizing flows},
  author={Rezende, Danilo Jimenez and Mohamed, Shakir},
  booktitle={International Conference on Machine Learning},
  pages={1530--1538},
  year={2015}
}

@inproceedings{dinh2017density,
  title={Density estimation using Real {NVP}},
  author={Dinh, Laurent and Sohl-Dickstein, Jascha and Bengio, Samy},
  booktitle={International Conference on Learning Representations},
  year={2017}
}

@article{xie2016cooperative,
  title={Cooperative training of descriptor and generator networks},
  author={Xie, Jianwen and Lu, Yang and Gao, Ruiqi and Zhu, Song-Chun and Wu, Ying Nian},
  journal={IEEE Transactions on Pattern Analysis and Machine Intelligence},
  volume={42},
  number={1},
  pages={27--45},
  year={2020},
  doi={10.1109/TPAMI.2018.2879081}
}

@inproceedings{han2017alternating,
  title={Alternating Back-Propagation for Generator Network},
  author={Han, Tian and Lu, Yang and Zhu, Song-Chun and Wu, Ying Nian},
  booktitle={Proceedings of the Thirty-First AAAI Conference on Artificial Intelligence},
  pages={1976--1984},
  year={2017},
  doi={10.1609/aaai.v31i1.10902}
}

@inproceedings{nijkamp2019anatomy,
  title={On the anatomy of {MCMC}-based maximum likelihood learning of energy-based models},
  author={Nijkamp, Erik and Hill, Mitch and Han, Tian and Zhu, Song-Chun and Wu, Ying Nian},
  booktitle={Proceedings of the AAAI Conference on Artificial Intelligence},
  volume={34},
  number={4},
  pages={5272--5280},
  year={2020}
}

\end{document}